\documentclass[12pt]{article}
\usepackage{times}
\usepackage{url}
\usepackage{graphicx}
\usepackage{algorithm}
\usepackage{algorithmic}
\usepackage{amsfonts}
\usepackage{amssymb}
\usepackage{amsmath}
\usepackage{bm}
\usepackage{scicite}

\newenvironment{sciabstract}{%
\begin{quote} \baselineskip14pt\small\hfil {\bf Abstract} \hfil\\[3pt]}
{\end{quote}\vspace{6pt}}

\newcounter{lastnote}

\title{On Reinforcement Learning for Full-length Game of StarCraft}

\author
{Zhen-Jia Pang, Ruo-Ze Liu, Zhou-Yu Meng, Yi Zhang, Yang Yu$^\dag$, Tong Lu\\
\\
\normalsize{National Key Laboratory for Novel Software Technology,}\\
\normalsize{Nanjing University, Nanjing 210023, China}\\
\normalsize{$^\dag$To whom correspondence should be addressed; E-mail: yuy@nju.edu.cn.}
}

\date{}

\begin{document}

\baselineskip16pt

% Make the title.

\maketitle 

\begin{sciabstract}%   <- trailing '%' for backward compatibility of .sty file
StarCraft II poses a grand challenge for reinforcement learning. The main difficulties include huge state space, varying action space, long horizon, etc. In this paper, we investigate a set of techniques of reinforcement learning for the full-length game of StarCraft II. We investigate a hierarchical approach, where the hierarchy involves two levels of abstraction. One is the macro-actions extracted from expert's demonstration trajectories, which can reduce the action space in an order of magnitude yet remains effective. The other is a two-layer hierarchical architecture, which is modular and easy to scale. We also investigate a curriculum transfer learning approach that trains the agent from the simplest opponent to harder ones. On a 64$\times$64 map and using restrictive units, we train the agent on a single machine with 4 GPUs and 48 CPU threads. We achieve a winning rate of more than 99\% against the difficulty level-1 built-in AI. Through the curriculum transfer learning algorithm and a mixture of combat model, we can achieve over 93\% winning rate against the most difficult non-cheating built-in AI (level-7) within days. We hope this study could shed some light on the future research of large-scale reinforcement learning.
\end{sciabstract}

\section{Introduction}
In recent years, reinforcement learning \cite{sutton1998introduction} (RL) has developed rapidly in many different domains. The game of Go has been considered conquered after AlphaGo \cite{silver2016mastering} and AlphaGo Zero \cite{silver2017mastering} . Many Atari games are nearly solved using DQN \cite{mnih2013playing} and follow-up methods. Various mechanical control problems, such as robotic arms \cite{levine2016end} and self-driving vehicles \cite{shalev2016safe}, have made great progress. However, reinforcement learning algorithms at present are still difficult to be used in large-scale reinforcement learning problem. Agents can not learn to solve problems as smartly and efficiently as humans. In order to improve the ability of reinforcement learning, complex strategic games like StarCraft have become the perfect simulation environments for many institutions such as DeepMind \cite{vinyals2017starcraft}, FAIR \cite{tian2017elf}, and Alibaba \cite{peng2017multiagent}.

From the perspective of reinforcement learning, StarCraft is a very difficult problem. Firstly, it is an imperfect information game. Players can only see a small area of map through a local camera and there is a fog of war in the game. Secondly, the state space and action space of StarCraft are huge. StarCraft's image size is much larger than that of Go. There are hundreds of units and buildings, and each of them has unique operations, making action space extremely large. Thirdly, a full-length game of StarCraft usually lasts from 30 minutes to more than an hour, and players need to make thousands of decisions to win. Finally, StarCraft is a multi-agent game. The combination of these issues makes StarCraft a great challenge for reinforcement learning.

Most previous agents in StarCraft are based on manual rules and scripts. Some works related to reinforcement learning are usually about micromanagement (e.g. \cite{usunier2016episodic}) and macromanagement (e.g. \cite{justesen2017learning}). These works solved some specific problems like local combat in StarCraft. However, there are rare works about the full-length games. In the paper of SC2LE \cite{vinyals2017starcraft}, the benchmark result given by DeepMind shows that the A3C algorithm \cite{mnih2016asynchronous} in SC2LE did not achieve one victory on the easiest level-1, which reveals the difficulty of the full-length game in StarCraft II. In the next section, we will have a summary of the difficulties encountered in the StarCraft II and introduce the simulation platform SC2LE (StarCraft II Learning Environment).

In this paper, we investigate a set of techniques of reinforcement learning for the full-length game of StarCraft II (SC2). In the section following the background, we present the hierarchical architecture investigated in this paper, which uses several levels of abstractions to make intractable large-scale reinforcement learning problems easier to handle. An effective training algorithm tailed to the architecture is also investigated. After that, we give some experiments in the full-length game on a 64x64 map of SC2LE. At last, we discuss about impacts of architectures, reward design and settings of curriculum learning. Experimental results achieved in several difficult levels of full-length games on SC2LE illustrate the effectiveness of our method. The main contributions of this paper are as follow:
\begin{itemize}
	\item We investigate a hierarchical architecture which makes large-scale SC2 problem easier to handle.
	\item A simple yet effective training algorithm for this architecture is also presented.
	\item We study in detail the impact of different training settings on our architecture.
	\item Experiment results on SC2LE show that our method achieves state-of-the-art results.
\end{itemize}

\section{Background}

\subsection{Reinforcement Learning}
Consider a finite-horizon Markov Decision Process (MDP), which can be specified as 6-tuple:
\begin{equation}
M=\langle S,A,P(.),R(.),\gamma, T \rangle,
\end{equation}
$ S $ is the state space and $ s \in S $ is a state of the state space. $ A $ is the action space and $ a \in A $ is an action which agent can choose in state $ s $. $ P(.)=\Pr(s' | s, a) $ represents the probability distribution of next state $ s' $ over $ S $ when agent choose action $ a $ in state $ s $. $ R(.)=R(s, a) $ represents the instant reward gained from the environment when agent choose action $ a $ in state $ s $. $ \gamma $ is discount factor which represents the influence of future reward on the choice at now. $ T $ is the max length of time horizon.

%If $ \pi(s) $ is a scalar and agent select an action $ a=\pi(s) $ when it is in state $ s $, this policy is called a distinct policy. In contrast, if $ \pi(s) $ is a distribution and agent samples an action $ a\sim\pi(s) $, this policy is called a stochastic policy.

Policy $ \pi $ is a mapping or distribution form $ S $ to $ A $. Assuming one agent, using a policy $ \pi $, starts from state $ s_0 $, chooses an action $ a_0 $,  gains a reward $ r_0=R(s_0,a_0) $, then transforms to next state $ s_1 $ according to distribution $ \Pr( .| s, a) $ and repeats this process. This will generate a sequence $ \tau $ below:
\begin{equation}
\tau = s_0, a_0, r_0, s_1, a_1, r_1,\ldots,
\label{eqn:d_i}
\end{equation}

There is a state $ s_{end} $ in which the agent will stop when the agent reaches this state. Process from $ s_0 $ to $ s_{end} $ is called one episode. The sequence $\tau$ in the episode is called a trajectory of the agent. For finite-horizon problem, when time step exceeds $ T $, the exploration is also over. A typical RL algorithm requires hundreds to thousands of episodes to learn a policy. In one episode, discounted cumulative reward get by the agent is defined as:
\begin{equation}
G = r_0 + \gamma r_1 +   \gamma^2 r_2 + \cdots,
\label{eqn:r_i}
\end{equation}
$ G $ is called return of the episode. A typical RL algorithm aims to find an optimal policy which maximizes the expected return.
\begin{equation}
\pi^* = \operatorname*{argmax}_{\pi}\mathbb{E}_{\pi}[\sum_{t=0}^{T} \gamma^t R(s_t,a_t)]
\end{equation}

\subsection{Hierarchical Reinforcement Learning}
When the dimension of the state space in the environment is huge, the space that needs to be explored exhibits exponential growth, which is called the curse of dimensionality problem in RL. Hierarchical reinforcement learning (HRL) solves this problem by decomposing a complex problem into several sub-problems and solving each sub-question in turn. There are some traditional HRL algorithms. Option \cite{sutton1999between} made abstraction for actions. MaxQ \cite{Dietterich1999maxq} split up the problem by the value function decomposition. And ALISP \cite{David2002alisp} provided a safe state abstraction method that maintains hierarchical optimality. Although these algorithms can better solve curse of dimensionality problems, they mostly need to be manually defined, which is time consuming and laborious. Another advantage of the HRL algorithm is that the resolution of the time is reduced, so that the problem of credit assignment over a long time scale can be better handled.

In recent years, some novel HRL algorithms have been proposed. Option-Critic \cite{bacon2017option} is a method using theorem of gradient descent to learn the options and the corresponding policies simultaneously, which reduces the effort of manual designing options. However, the automatically learned options do not perform as well as non-hierarchical algorithms on certain tasks. FeUdalNetwork \cite{vezhnevets2017feudal} designed a hierarchical architecture that includes a Manager module and a Worker module and propose a gradient transfer strategy to learn the parameters of the Manager and Worker in an end-to-end manner. However, due to the complexity, this architecture is hard-to-tune. MLSH \cite{frans2017meta} proposes a hierarchical learning approach based on meta-learning, which enhances the learning ability of transferring to new tasks through sub-policies learned in multiple tasks. MLSH has achieved better results on some tasks than the PPO \cite{schulman2017proximal} algorithm, but because its setting is multi-tasking, it is difficult to apply to our environment.

%The hierarchical framework we propose in this paper consists of two levels of abstraction, one is architectural abstraction and the other is abstraction of action space. We use the mechanism of data mining to automatically learn the combined macro actions from the expert's trajectory. Not only does it save the effort of manual design, but also ensures the effectiveness of macro actions. In terms of architecture design, we refer to the design of FeUdalNetwork and design the top layer network controller and the lower layer network Executer. For the sake of simplicity, we draw on the ideas in MLSH to design our training algorithm. Through two layers of abstraction, our framework can learn an effective strategy more efficiently. Another advantage of our hierarchical framework is modular and easy to scale, which will be discussed later.

\subsection{StarCraft II}
%Games are ideal environments for reinforcement learning research. RL problems on real-time strategy (RTS) games are far more difficult than problems on Go due to complexity of states, diversity of actions, and long time horizon. After AlphaGo, Research on AI is more focused on RTS games.

Games are ideal environments for reinforcement learning research. RL problems on real-time strategy (RTS) games are far more difficult than problems on Go due to complexity of states, diversity of actions, and long time horizon. Traditionally, research on real-time strategy games is based on search and planning approaches \cite{ontanon2013survey}. In recent years, more studies use RL algorithms to conduct research on RTS and one of the most famous RTS research environments is StarCraft. Previous works on StarCraft are mostly focused on local battles or part-length and often get features directly from game engine. \cite{usunier2016episodic} presents a heuristic reinforcement learning algorithm combining exploration in the space of policy and back propagation. \cite{peng2017multiagent} introduces BiCNet based on multi-agent reinforcement learning combined with actor-critic. Although they have achieved good results, they are only effective for part-length game. ELF \cite{tian2017elf} provides a framework for efficient learning and a platform mini-RTS for reinforcement learning research. ELF also gives a baseline of A3C \cite{mnih2016asynchronous} algorithm in a full-length game of mini-RTS. However, because the problem is relatively simple, there is still a great distance from the complexity of StarCraft.

SC2LE is a new research learning environment based on StarCraft II which is the follow-up of StarCraft. The location information for each unit is given in the engine of StarCraft. However, the location information of units and buildings need to be perceived from the image features in SC2LE. Therefore, the spatial complexity of the state of its input is much larger than StarCraft I. At the same time, in order to simulate real hand movement of humans, action space in SC2LE is refined to each mouse click event, which greatly increases the difficulty of learning and searching. Benchmark result given by DeepMind shows that A3C algorithm \cite{mnih2016asynchronous} did not achieve one victory on the easiest difficulty level-1, verifying the difficulties of full-length game in StarCraft II. In addition to full-length game, SC2LE also provides several mini-games for research. \cite{zambaldi2018relational} proposed a relation-based reinforcement learning algorithm, which achieved good results on these mini-games. But the results on the full-length game are still not reported in this paper. Recently, there is a paper \cite{Sun2018tsbot} about the full-length game of StarCraft II, which achieves good results. Our performance is close to their work, but we use less prior knowledge and effectively exploit image information. Moreover, we use less computing resources.

\section{Methodology}
%StarCraft is a large scale reinforcement learning problem, and it's hard to learn a single policy to win the game. From the hierarchical perspective, we usually need to achieve multiple goals to get the final victory. For example, we need to earn our money as much as possible, beat the enemy really hard in the battle and so on.  Instead of learning the single policy, we can split a large problem into several small pieces. Every goal in the game can be seen as a small reinforcement learning problem and can be learned by a sub-policy. At the same time, there is a high level controller to learn how to use all these sub-policy.

In this section, we introduce our hierarchical architecture and the generation of macro-actions firstly. Then the training algorithm of the architecture is given. At last, we discuss the reward design and the curriculum learning setting used in our method.

\begin{figure}[h!]
	\begin{minipage}[t]{\linewidth}
		\centering
		\includegraphics[width=\textwidth]{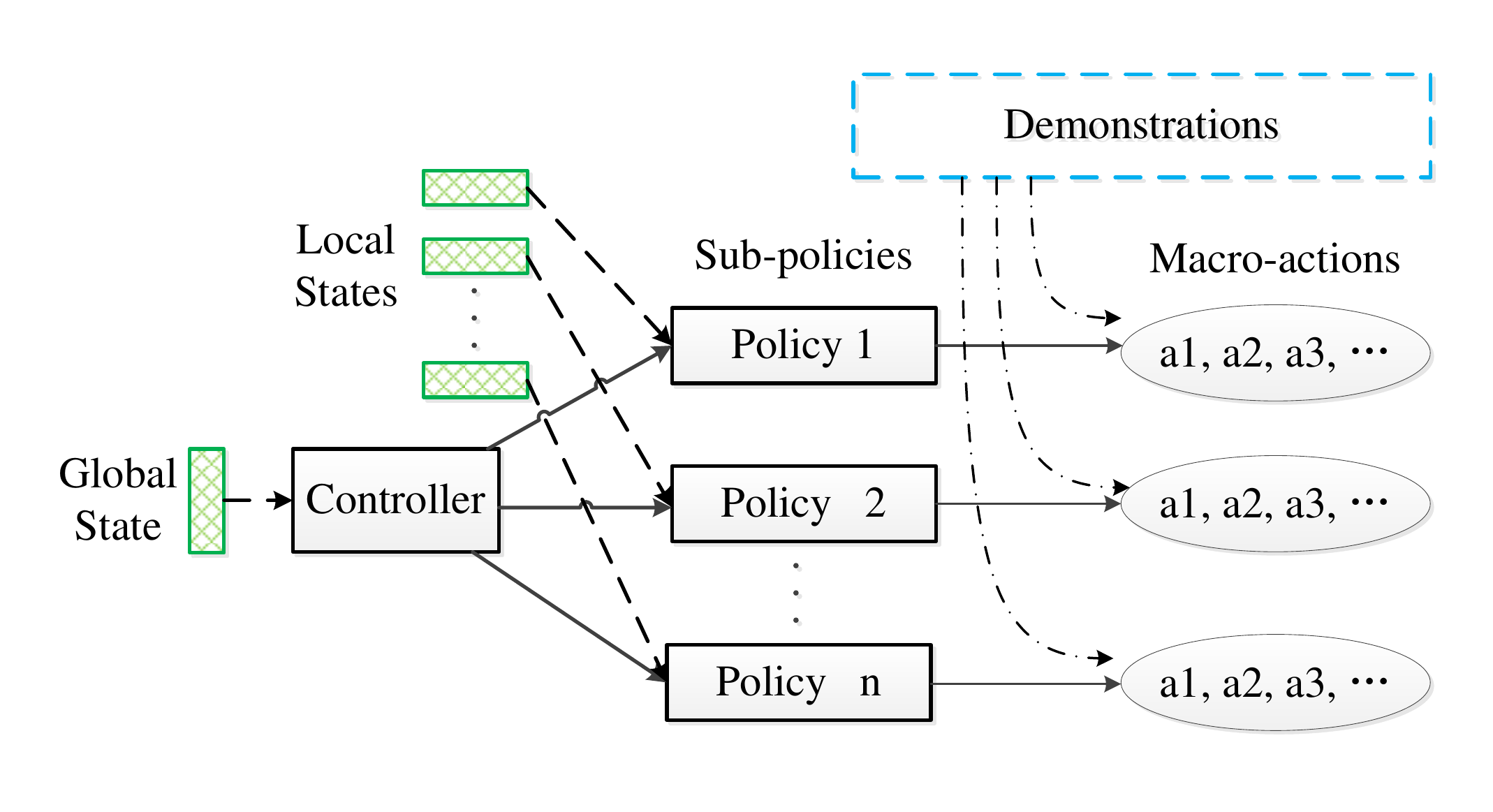}
		\caption{Overall Architecture.}
		\label{fig:Arch}
	\end{minipage}
\end{figure}

\subsection{Hierarchical Architecture}
Our hierarchical architecture is illustrated in Fig. \ref{fig:Arch}. There are two types of policies running in different timescales. The controller decides to choose a sub-policy based on current observation every long time interval, and the sub-policy picks a macro-action every short time interval.

%\subsubsection{Controller and sub-policy pool}
For further illustration, we use $\Pi$ to represent the controller. $S_c$ and $A_c$ is its state and action space. %$R_c$ is the reward function of controller.
Similarly, assuming there is $n$ sub-policies in the pool, we use $\langle \pi_1, \pi_2, ..., \pi_n \rangle$ to represent them. The state and action space of $i$th sub-policy is defined as $S_i$ and $A_i$. $R_i$ is its reward function. Besides, we have a time interval K. It means that the controller chooses a sub-policy in every K time units and the chosen sub-policy makes a decision in every time unit. Now we can deeply go through the whole process.

At time $t_c$, the controller gets its own global observation $s^c_{t_c}$, and it will choose a sub-policy $i$ based on its state, like below:
\begin{equation}
a_{t_c}^c =\Pi(s^c_{t_c}),  \quad  s^c_{t_c} \in S_c
\end{equation}

Now the controller will wait for K time units and the $i$th sub-policy begins to make its move. We assume its current time is $t_i$ and its local observation is $s^i_{t_i}$,  so it get the macro-action $a^i_{t_i} = \pi_i(s^i_{t_i})$. After the $i$th sub-policy doing the macro-action $a^i_{t_i}$ in the game, it will get the reward $r^i_{t_i} = R_i(s^i_{t_i}, a^i_{t_i})$ and its next local observation  $s^i_{t_i + 1}$. The tuple $ (s^i_{t_i}, a^i_{t_i}, r^i_{t_i}, s^i_{t_i + 1})$ will be stored in its local buffer $D_i$ for the future training. After K moves, it will return to the controller and wait for the next chance.

The high level controller gets the return of the chosen sub-policy $\pi_i$ and compute the reward of its action $a_{t_c}^c$ as follows:
\begin{equation}
r_{t_c}^c = r^i_{t_i} + r^i_{t_i+1} + ... + r^i_{t_i+K-1}
\end{equation}

Also,  the controller will get its next global state $s^c_{t_c+1}$ and the tuple $ (s^c_{t_c}, a^c_{t_c}, r^c_{t_c}, s^c_{t_c + 1})$ will be stored in its local buffer $D_c$. Now the time is $t_c + 1$, and the controller will make a next choice based on its current global observation.

From above, we can see that there is some advantages in our hierarchical architecture. Firstly, each sub-policy and the high-level controller have different state space. We can see that the controller only needs the global information to make high-level decision. The global state space $S_c$ is a small part of all the state space $S$. Also, a sub-policy responsible for combat is more focused on local state space $S_i$ related to battle. It can be seen that such a hierarchical structure can split the original huge state space into a plurality of subspaces corresponding to different policy networks. Secondly, the hierarchical structure can also split the tremendous action space $A$. The sub-policies with different functions will have their own action space $A_i$. Thirdly, the hierarchical architecture can effectively reduce the execution step size of the strategy. Since the control network calls a sub-network every fixed time interval $K$ , the total execution step size of the high-level network becomes $T/K$ step. The execution step size of sub-policies will also be reduced. Last but not least, the hierarchical architecture makes design of the reward functions easier. Different sub-policies may have different targets. Therefore, they can learn more quickly by their own suitable reward functions.

\subsection{Generation of Macro-actions}
In StarCraft, the original action space $A$ is tremendous. And human players always need to do a sequence of raw actions to achieve one simple purpose. For example, if we want to build a building in the game, we have to select a peasant, order it to build the building in the specific position, and make it come back after finish. The sequences of the raw actions for some simple purposes are more likely some fixed sequence stored in mind for our human players. So we instead generate a macro-action space $A^{\eta}$ which is obtained through data mining from trajectories of experts. The original action space $A$ is then replaced by the macro-action space $A^{\eta}$. This will improve the learning efficiency and testing speed. The generation process of macro-actions is as follow:

\begin{itemize}
	\item Firstly, we collect some expert trajectories which are sequence of operations $ a \in A $ from game replays.
	\item Secondly, we use a prefix-span \cite{yan2003clospan} algorithm to mine the relationship of the each operation and combine the related operations to be a sequence of actions $a^{seq}$ of which max length is $C$ and constructed a set $A^{seq}$ which is defined as $ A^{seq} =\{\, a^{seq} = (a_0, a_1, a_2, ... a_i) \mid a_i \in A \text{ and } i \leq C, \,\} $
	\item Thirdly, we sort this set by $\operatorname*{frequency}(a^{seq})$.
	\item Fourthly, we remove duplicated and meaningless ones, remain the top $K$ ones. Meaningless refers to the sequences like continuous selection or camera’s movement.
	\item Finally, the reduced set are marked as newly generated macro-action space $A^{\eta}$.
\end{itemize}

Using the macro-action space $A^{\eta}$, our MDP problem is now reduced to a simple one, which is defined as:
\begin{equation}
M =\langle S,A^{\eta}, P(.), R(.), \gamma, T \rangle,
\end{equation}

Meanwhile, the MDP problem of each sub-policy is also reduced to new one:
\begin{equation}
M_{i} =\langle S_{i}, A_{i}^{\eta}, P(.), R_i(.), \gamma, T_{i} \rangle, \text{ for } i = 1 \text{ to } n,
\end{equation}

\begin{algorithm}[H]
	\caption{HRL training algorithm} %
	{\bf Input:} Number of sub-policies $N$, time interval K, reward function $R_1, R_2, ..., R_n$, max episodes $M$, max iteration steps $Z$, max game steps $T$,
	\begin{algorithmic}
		\STATE Initialize replay buffer $\langle D_c, D_1, D_2, ..., D_n\rangle$ , controller policy $\Pi_\phi$, each sub-policy $\pi_{\theta_i}$
		\FOR{$z=1$ to $Z$}
		\STATE clear $\langle D_c, D_1, D_2, ..., D_n\rangle$
		\FOR{$m=1$ to $M$}
		\STATE clear $\langle \tau_c, \tau_1, \tau_2, ..., \tau_n\rangle$
		\FOR{$t=0$ to $T$}
		\IF {$t$ mod $K == 0$}
		\IF {$s^c_{t_c-1}$ is not None}
		\STATE $\tau_c \gets \tau_c \cup \{(s^c_{t_c-1}, a^c_{t_c-1}, r^c_{t_c-1}, s^c_{t_c})\}$
		\ENDIF
		\STATE $a^c_{t_c} \gets \Pi_\phi(s^c_{t_c}), \quad r^c_{t_c} \gets 0$
		\STATE $j \gets a^c_{t_c}$
		\ENDIF
		\STATE collect experience by using $\pi_{\theta_j}$
		\STATE $\tau_j \gets \tau_j  \cup \{(s_{t}^j, a_{t}^j, R_j(s_{t}^j, a_{t}^j), s_{t+1}^j\}$
		\STATE $r^c_{t_c} \gets R_j(s_{t}^j, a_{t}^j) + r^c_{t_c} $
		\ENDFOR
		\STATE $D_c \gets D_c \cup \tau_c$
		\FOR{$i=1$ to $N$}
		\STATE $D_i \gets D_i \cup \tau_i$
		\ENDFOR
		\ENDFOR
		\STATE using $D_c$ to update $\phi$ to maximize expected return % from $1/K$ timescale viewpoint
		\FOR{$i=1$ to $N$}
		\STATE using $D_i$ to update $\theta_i$ to maximize expected return % from full timescale viewpoint
		\ENDFOR
		\ENDFOR
	\end{algorithmic}
	\label{alg: train}
\end{algorithm}

\subsection{Training Algorithm}
The training process of our architecture is showed in Algorithm \ref{alg: train} and can be summarized as follows. Firstly we initialize the controller and sub-policies. Then we run the iteration of $Z$ times and run the episode of $M$ times in each iteration. At the beginning of each iteration, we will clear all the replay buffers. In each episode, we collect the trajectories of the controller and sub-policies. At the end of each iteration, we use the replay buffers to update the parameters of the controller and sub-policies.

The update algorithm we use is PPO \cite{schulman2017proximal}. Entropy's loss was added to the PPO's loss calculation to encourage exploration. Therefore, our loss formula is as follows:
\begin{equation}
L_t(\theta) = \hat{\mathbb{E}}_t[L_t^\text{clip}(\theta) + c_1 L_t^\text{vf}(\theta) + c_2 S[\pi_\theta](s_t) ]
\end{equation}
where $c_1, c_2$ are the coefficients we need to tune, and S denotes an entropy bonus. $L_t^\text{clip}(\theta)$ is defined as follows:
\begin{equation}
L_t^\text{clip}(\theta) = \hat{\mathbb{E}}_{t}[\text{min}(r_t(\theta)\hat{A}_t, \text{clip}(r(\theta), 1-\epsilon, 1+\epsilon)\hat{A}_t]
\end{equation}
\begin{equation}
L_t^\text{vf}(\theta) = \hat{\mathbb{E}}_{t}[ (r(s_t, a_t) + \hat{V}_t(s_t) - \hat{V}_t(s_{t+1}))^2 ]
\end{equation}
where $r_t(\theta)= \frac{\pi_\theta(a_t|s_t)}{\pi_{\theta_{old}}(a_t|s_t)}$, $\hat{A}_t$ is computed by a truncated version of generalized advantage estimation.

\subsection{Reward Design}
%As we know, reward has a significant impact on reinforcement learning. There are usually two types of reward we can use. One is dense reward that can be got easily during the game, and the other is sparse reward that only shown at few specific states. Obviously, dense reward gives more positive or negative feedback to the agent. As a result, dense reward can help agent learn to play the game faster and better than sparse reward.

There are three types of rewards we explore in this paper. Win/Loss reward is a ternary 1 (win) / 0 (tie) / -1 (loss) received at the end of a game. Score reward is Blizzard score get from the game engine. Mixture reward is our designed reward function. 

It's hard for agent to learn the game using Win/Loss reward. Blizzard scores can be seen as dense reward. We will show that using this score as a reward can not help agent get more chances to win in the experiment section. We have designed reward functions for the sub-policies which combines dense reward and sparse reward. These rewards seem to be really effective for training, which will be shown in the experiment section.

%This reward draws on the fact that human beings not only examine the results in their studies, but also pay attention to performance in the process, similar to the score calculation of courses in university.

%Our reward design is as follows:
%\begin{itemize}
%    \item 1. We collected replays of numerous games from the experts.
%    \item 2. We count the average number of units and buildings built in the replay by the experts.
%    \item 3. In the game, we calculate the difference between the number of units and buildings in each of the two adjacent states.
%    \item 4. For each unit or building, we calculate the reward based on this rules: if it is less than the expert's value, add a fixed reward, otherwise reduce a reward.
%    \item 5. At the end of the game, if the episode lasts for $M$ minutes, give the last state a reward penalty $ a * M $ .
%    \item 6. Give the last state a result reward which is multiplying the outcome of the game by a weight. This weight is the average number of the process reward of a lot of winning games.
%\end{itemize}

\subsection{Curriculum Learning}
%Curriculum learning is an effective method that can be applied to reinforcement learning. It designs a curriculum from easy to hard. Agents can continue to learn from this curriculum to improve their abilities. We use the idea of curriculum learning here to let our agents can continue to challenge more difficult tasks.

SC2 includes 10 difficult levels of built-in AI which are all crafted by rules and scripts. From level-1 to level-10, the built-in AI's ability is constantly improving. Training in higher difficulty level gives less positive feedback, making it difficult for agent to learn from scratch. In this work, we trained our agent in lower difficulty level at first, then transferred the agent to higher difficulty level using the pre-trained model as the initial model, following the idea of curriculum learning.

%Starting from level-3, the level of the built-in AI is close to that of ordinary human players. Starting with level-8, built-in AI begins to use cheating techniques. The most difficult three ones (8, 9, 10) using cheat to get more resources and vision, which are very difficult to beat.

However, when the pre-trained agent transfers to high difficulty levels, we find that if controller and all sub-policies are still updated at the same time, the training is sometimes unstable due to the mutual influence of different networks. In response to this situation, we have devised a strategy to update the controller and sub-policies alternatively. We found that this method can make the training more stable, and the winning rate for high difficulty levels can rise steadily.

\section{Experiments}
In this section, we present the experiment results on SC2LE. We first introduce the experiment settings including the hierarchical architecture and three combat models. Then the details of experiments are shown. If you want to get more information about experiment settings and implementation details such as features of state space or forms of macro actions, please refer to the appendix of our arxiv version \footnote{https://arxiv.org/abs/1809.09095}.

%At last, we discuss the results and analyze the possible causes.

\subsection{Setting}
The setting of our architecture is as follow: controller selects one sub-policy every 8 seconds, and the sub-policy performs macro-actions every 1 second. In the setup, we have two sub-policies in the sub-policy pool. One sub-policy controls the construction of buildings and the production of units in the base, called base network. The other sub-policy is responsible for battle, called battle policy. This sub-policy has three different models, which are explained later.

\begin{figure}[h!]
	\begin{minipage}[t]{\linewidth}
		\centering
		\includegraphics[width=\textwidth]{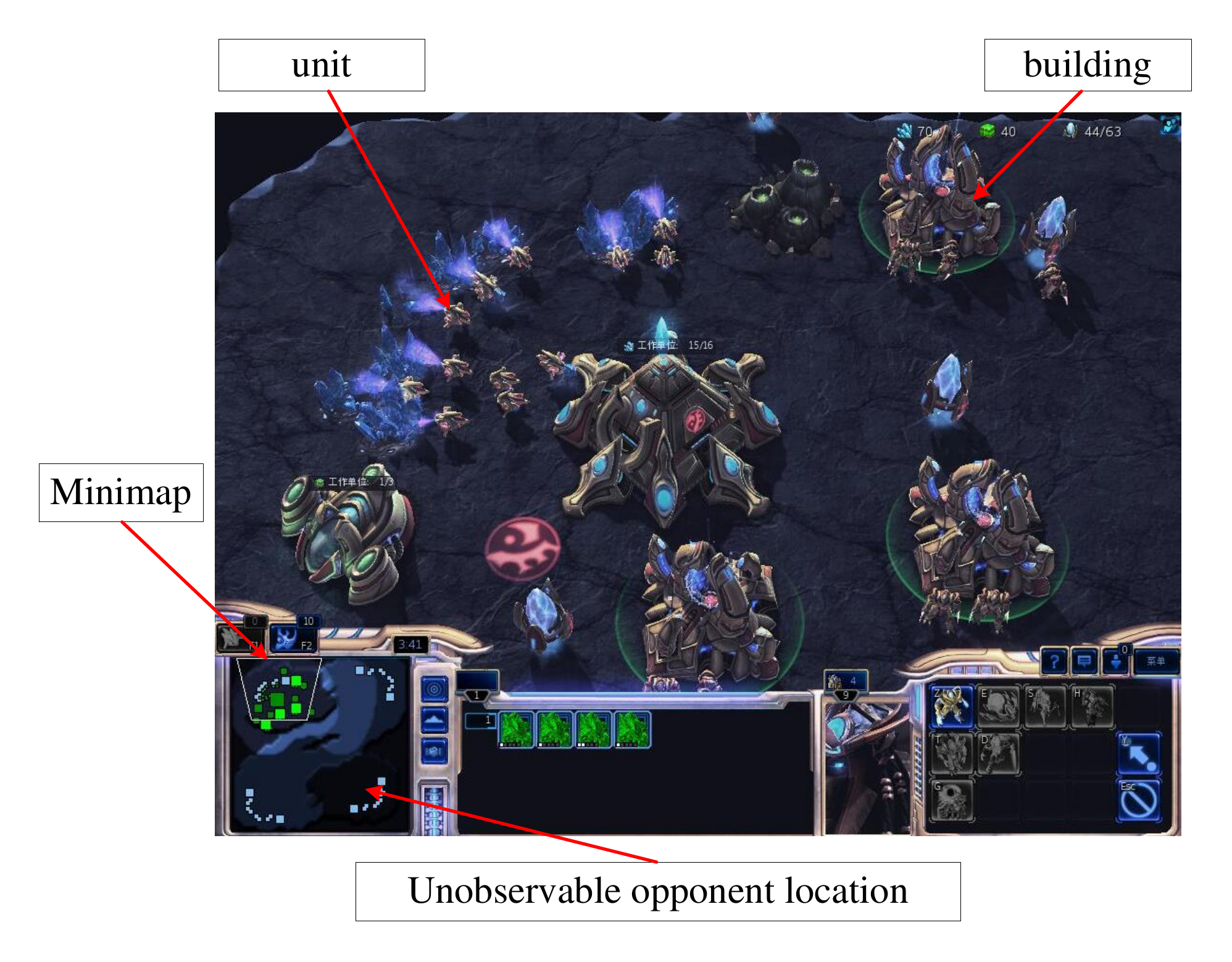}
		\caption{Screenshot of StarCraft II.}
		\label{fig:pic}
	\end{minipage}
\end{figure}

A full-length game of SC2 in the 1v1 mode is as follow: First, two players spawn on different random points in the map and accumulate resources. Second,  they construct buildings and produce units. At last, they attack and destroy each other's units and buildings. Fig. \ref{fig:pic} shows a screenshot of the running game.

For simplicity, we set our agent's race to \textit{Protoss} and build-in AI's race to \textit{Terran}. But our algorithms could generalize to any race against any other race. The map we used is the 64x64 map \textit{simple64} on SC2LE. We set the maximum length of each game to $15$ minutes. Moreover, our agent does not open sub-mine, and only uses the two basic military units which are \textit{Zealot} and \textit{Stalker}.

%We use a machine with 4 GPUs and 48 CPUs for the experiment. There are 10 workers and each worker has 5 threads. In other words, we has 50 running environments at the same time.  They all share the global hierarchical architecture and also the parameters of networks. Since we use PPO algorithm for each network, our distributed system is simultaneous. In each iteration, we make the workers run 100 full-length games of SC2 and each worker would collect the data from 10 episodes. It would takes about 4 minutes. When all the workers have collected the data, each worker would compute the gradients for each networks. All the gradients from the wokers would be gathered and used to update the global hierarchical architecture.

In our setup, the battle policy has three different settings. These are explained below.

\subsubsection{Combat rule}
Combat rule is a simple battle policy, and there is only one action in combat rule model: attacking a sequence of fixed positions. Although the map is unobservable and the enemy's position is unknown, the enemy always live around the mines. We only need to make our army attack the fixed positions around the mines. The attack action uses the built-in AI to do automatic move and attack. Thus the result of the attack action depends on the construction of the buildings and the production of the units. Only when the agent learns to better carry out building construction (for example, do not build redundant buildings, and build more pylons in time when supply is insufficient), it is possible for the agent to win.

\subsubsection{Combat network}
Though the simple combat rule model is effective, the combat process is slightly naive or rigid and may fail when moving to larger and more complex maps. Below we introduce a smarter attack approach which is called combat network. The output of the combat network consists of three actions and a position vector. These actions are: all attack a certain position, all retreat to a certain position, do nothing. Attack and move positions are specified by the position vector.

%The location position is specified by a position vector. The position vector is a one-hot vector that represents the eight coordinate points on the screen. We can imagine that there is a square in the center of the screen. The side length of the square is half the length of the side of the screen. These 8 points are evenly distributed on the side of rectangle, referring to the position of the attack and movement. With this setup, the agent can be smarter to choose the target and location of the offense, thus increasing the flexibility of the battle. At the same time, because the combat network does not specify the location of the enemy, it can automatically learn and discover the possible existence of the enemy, so that it can maintain performance when moving to a larger and more complex map.

%\begin{figure}[h!]
%	\begin{minipage}[t]{\linewidth}
%		\centering
%		\includegraphics[width=\textwidth]{./figure/Combatnet.pdf}
%		\caption{Architecture of combat network.}
%		\label{fig:combatnet}
%	\end{minipage}
%\end{figure}

% which is shown in Fig. \ref{fig:combatnet}
The combat network is constructed as a Convolution Neural Network (CNN). This CNN accepts feature maps of minimaps and screens which enabling it to know the information of the full map and the unit and building positions on the screen. Moreover, we use a simple strategy to decide the position of the camera. When the controller chooses the base sub-policy, the camera is moved to the location of agent's base. When the controller chooses the battle sub-policy, the camera is moved to the location of army. The location of army can be chosen in two ways. The first is the center point of all combat units. The second is the center of the most injured unit. Since the injured unit indicated the occurrence of a recent battle, we found that the last setting was better in practice.

\subsubsection{Mixture model}
Although the combat network model can be trained well on high level of difficulty, it will occasionally miss some hidden enemy buildings. We can combine combat network with combat rule into a mixture model. When a certain value is predicted in the position vector of the combat network, the army's attack position will become a series of fixed positions got by prior knowledge.

\begin{figure*}[t]
	\begin{minipage}{0.49\linewidth}
		\centering
		\includegraphics[width=\textwidth]{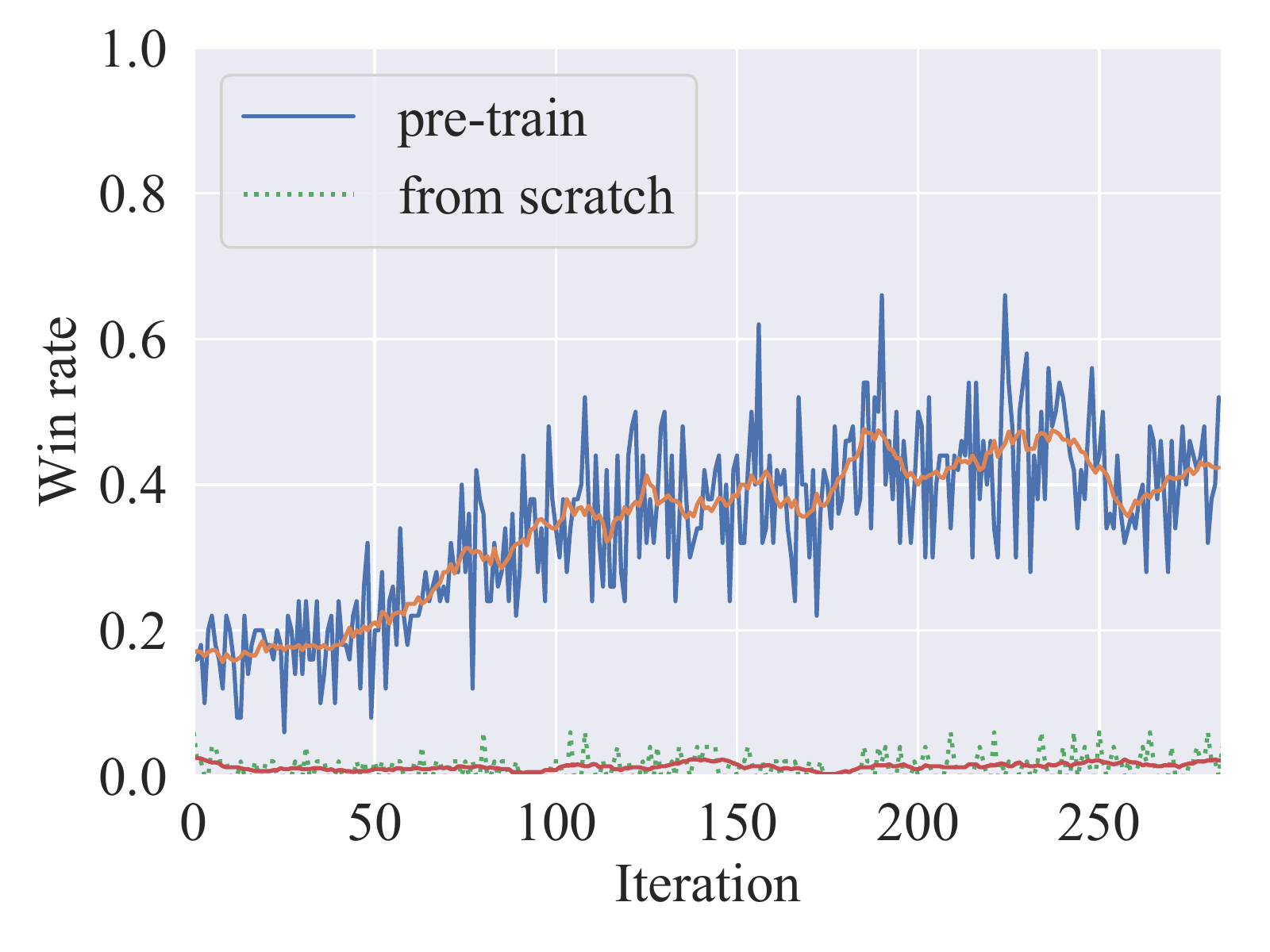}\\
		(a) Curriculum learning
		\label{fig:s.2.1.1}
	\end{minipage}%
	\begin{minipage}{0.49 \linewidth}
		\centering
		\includegraphics[width=\textwidth]{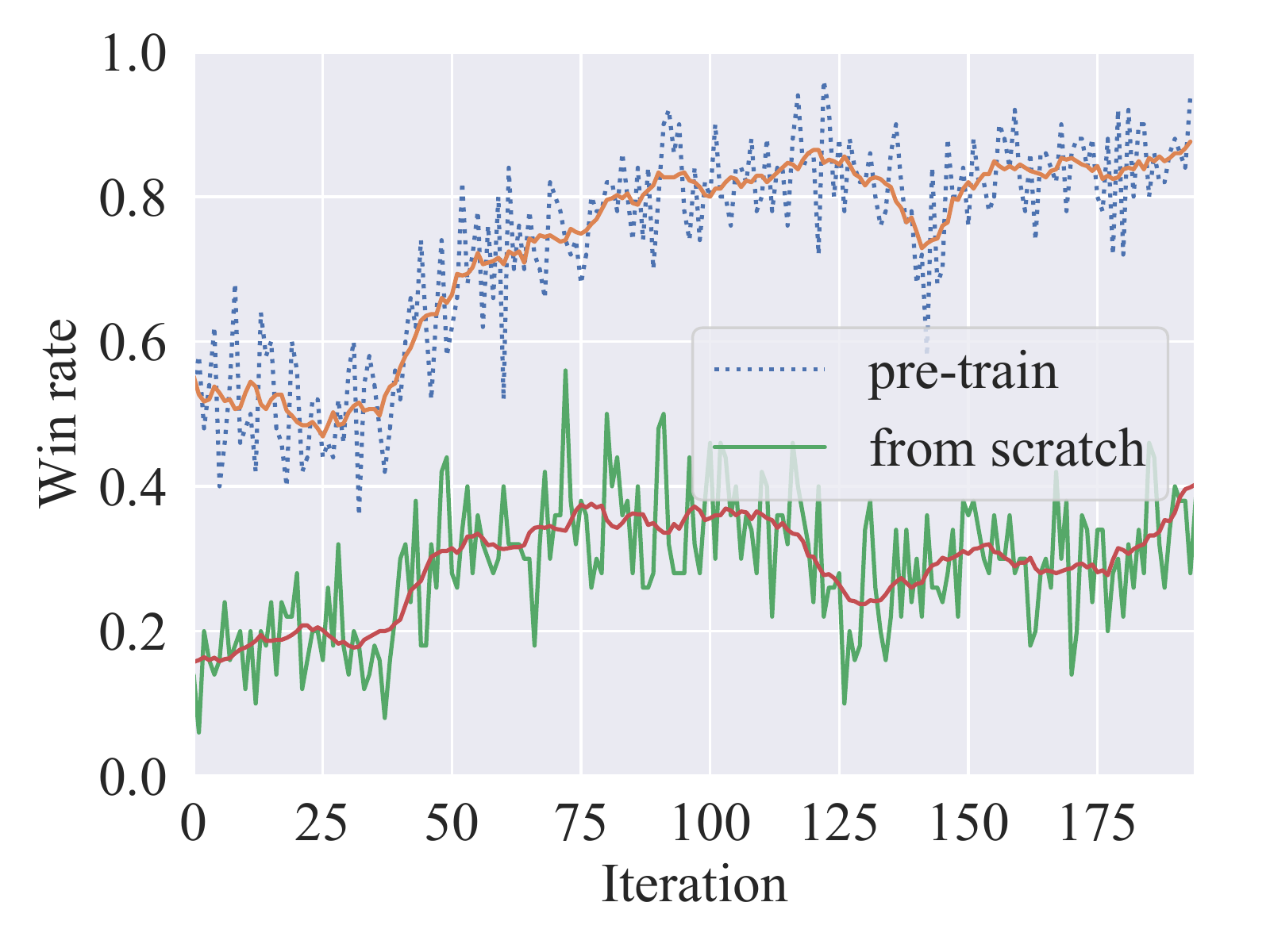}\\
		(b) Module training
		\label{fig:s.2.1.2}
	\end{minipage}
	\begin{minipage}{0.49 \linewidth}
		\centering
		\includegraphics[width=\textwidth]{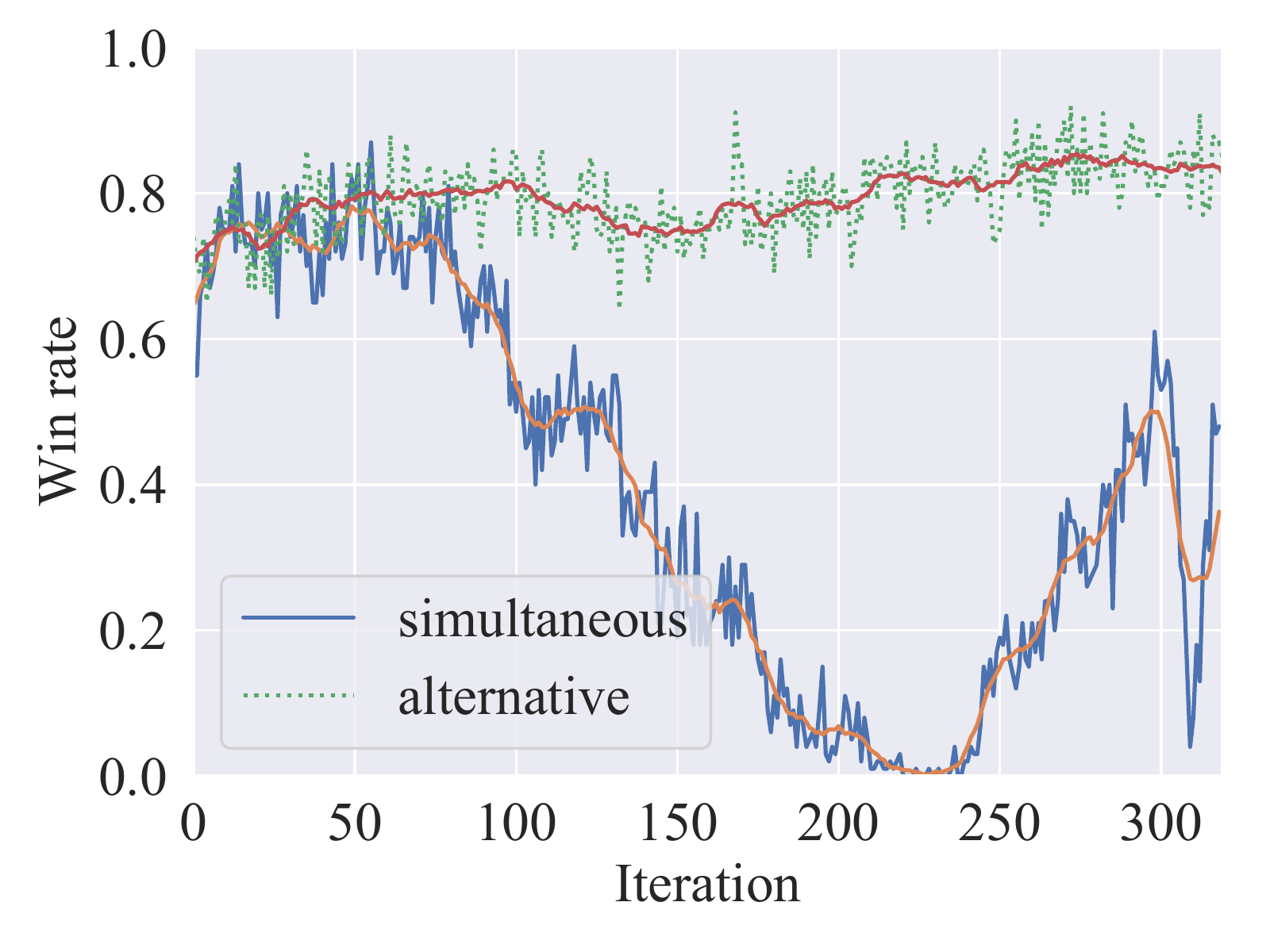}\\
		(c) Simultaneous unstable
		\label{fig:s.2.2.1}
	\end{minipage}
	\begin{minipage}{0.49 \linewidth}
		\centering
		\includegraphics[width=\textwidth]{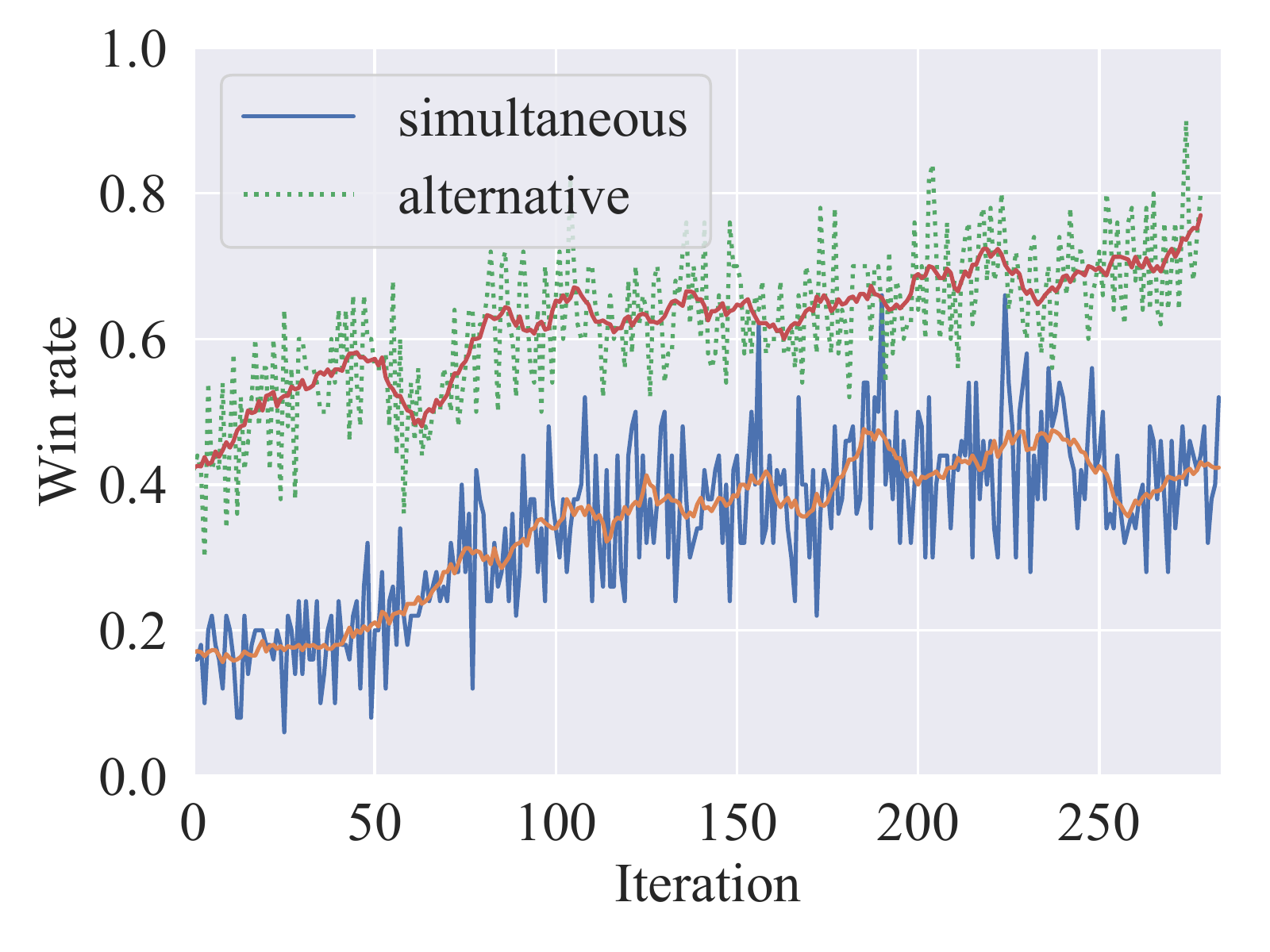}\\
		(d) Alternative refine
		\label{fig:s.2.2.2}
	\end{minipage}
	\caption{Winning curve in training process.}
	\label{fig:s.a.2}
\end{figure*}

\begin{figure*}[t]
	\begin{minipage}{0.49\linewidth}
		\centering
		\includegraphics[width=\textwidth]{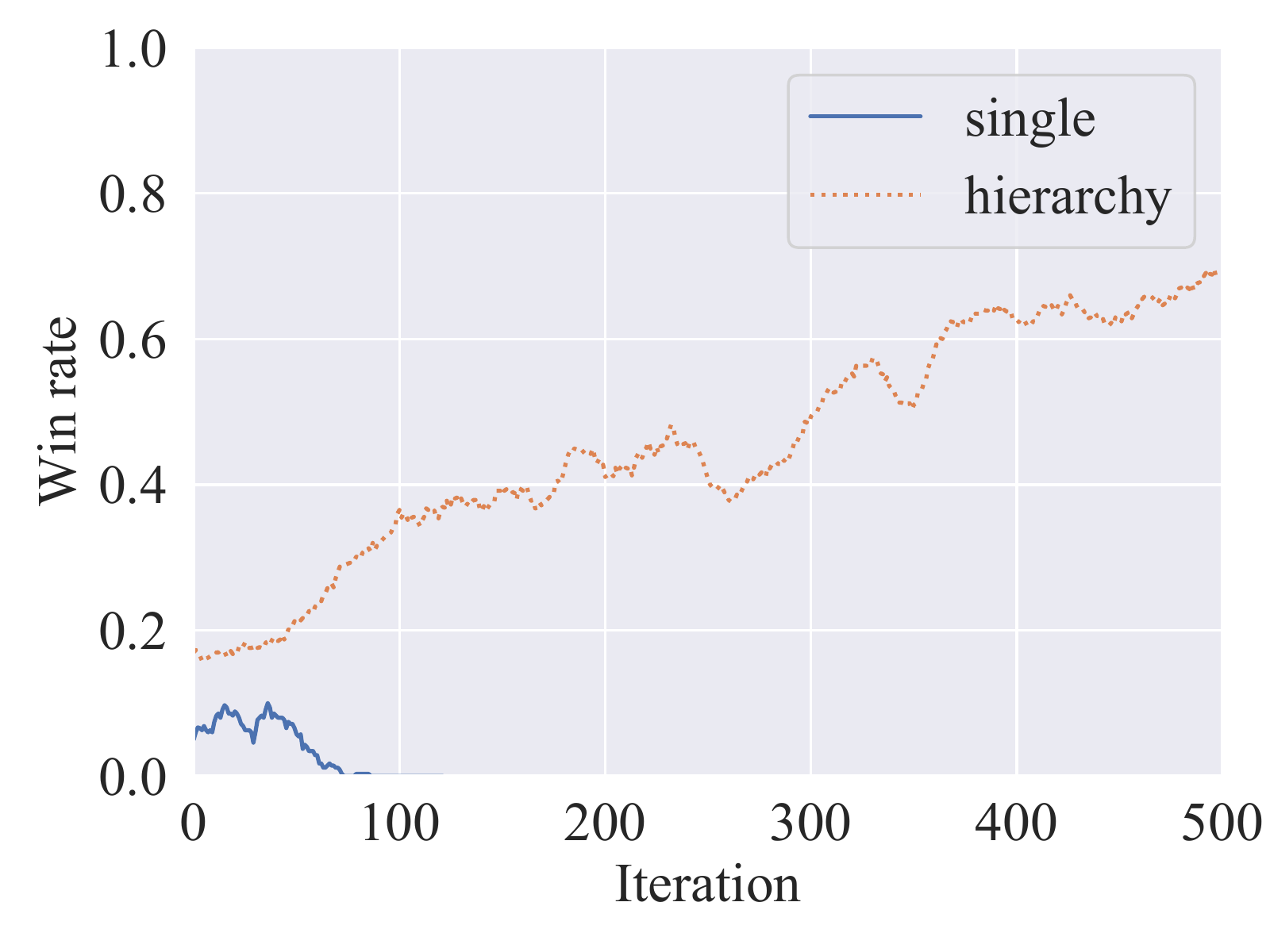}\\
		(a) Hierarchy vs single
		\label{fig:s.3.1.1}
	\end{minipage}%
	\begin{minipage}{0.49 \linewidth}
		\centering
		\includegraphics[width=\textwidth]{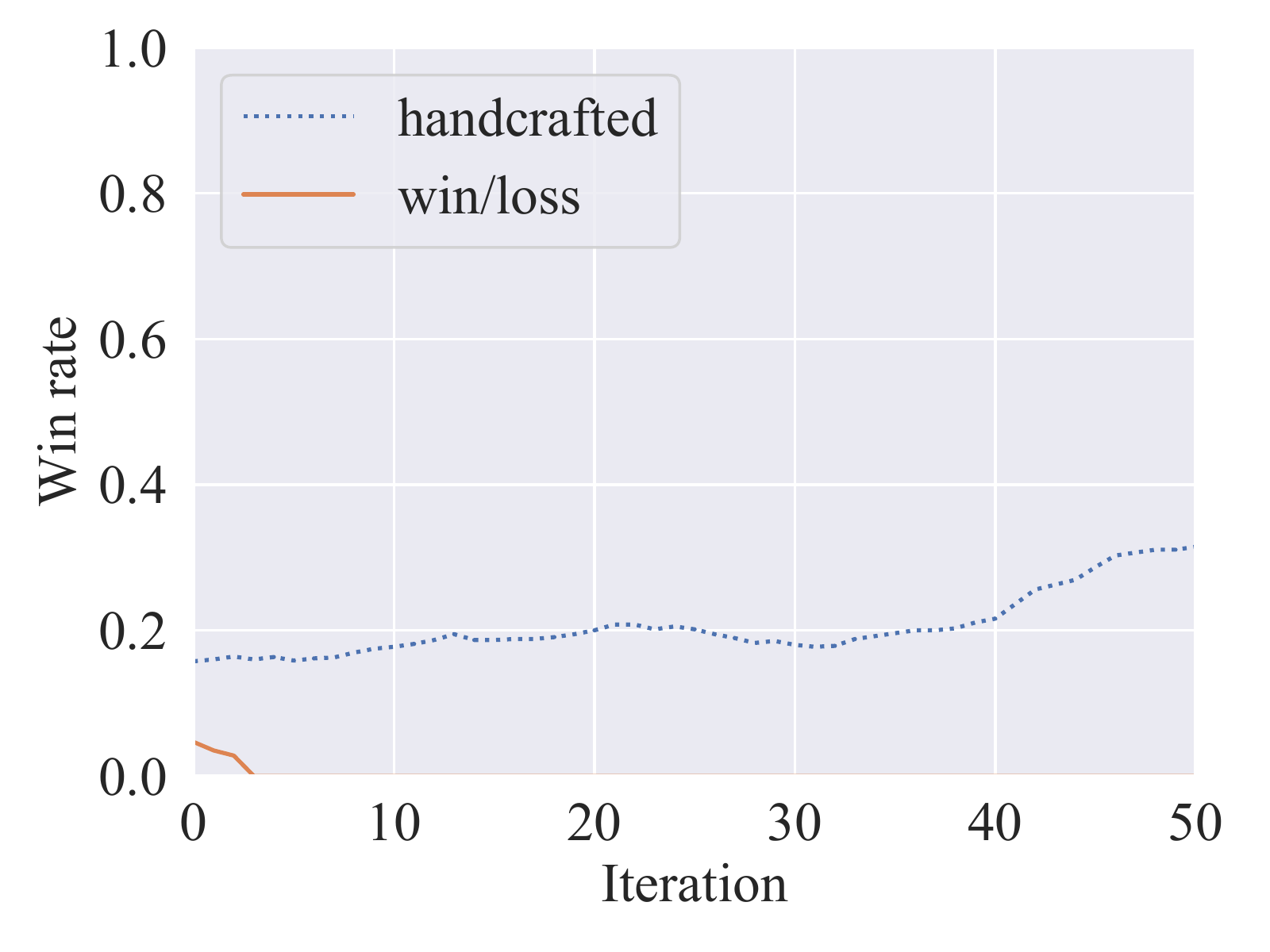}\\
		(b) Win/Loss vs handcrafted
		\label{fig:s.3.1.2}
	\end{minipage}
	\begin{minipage}{0.49 \linewidth}
		\centering
		\includegraphics[width=\textwidth]{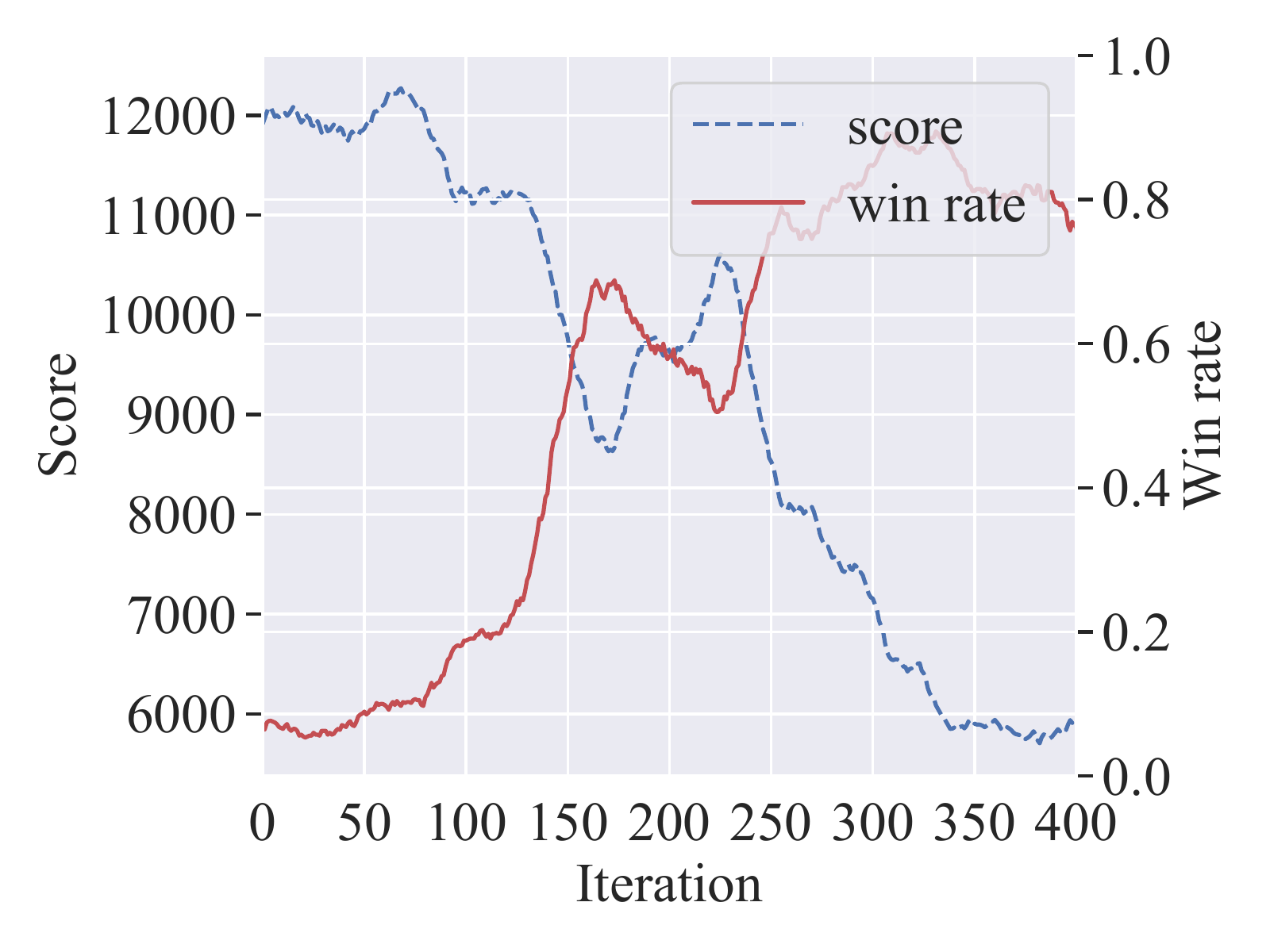}\\
		(c) Score and win rate
		\label{fig:s.3.2.1}
	\end{minipage}
	\begin{minipage}{0.49 \linewidth}
		\centering
		\includegraphics[width=\textwidth]{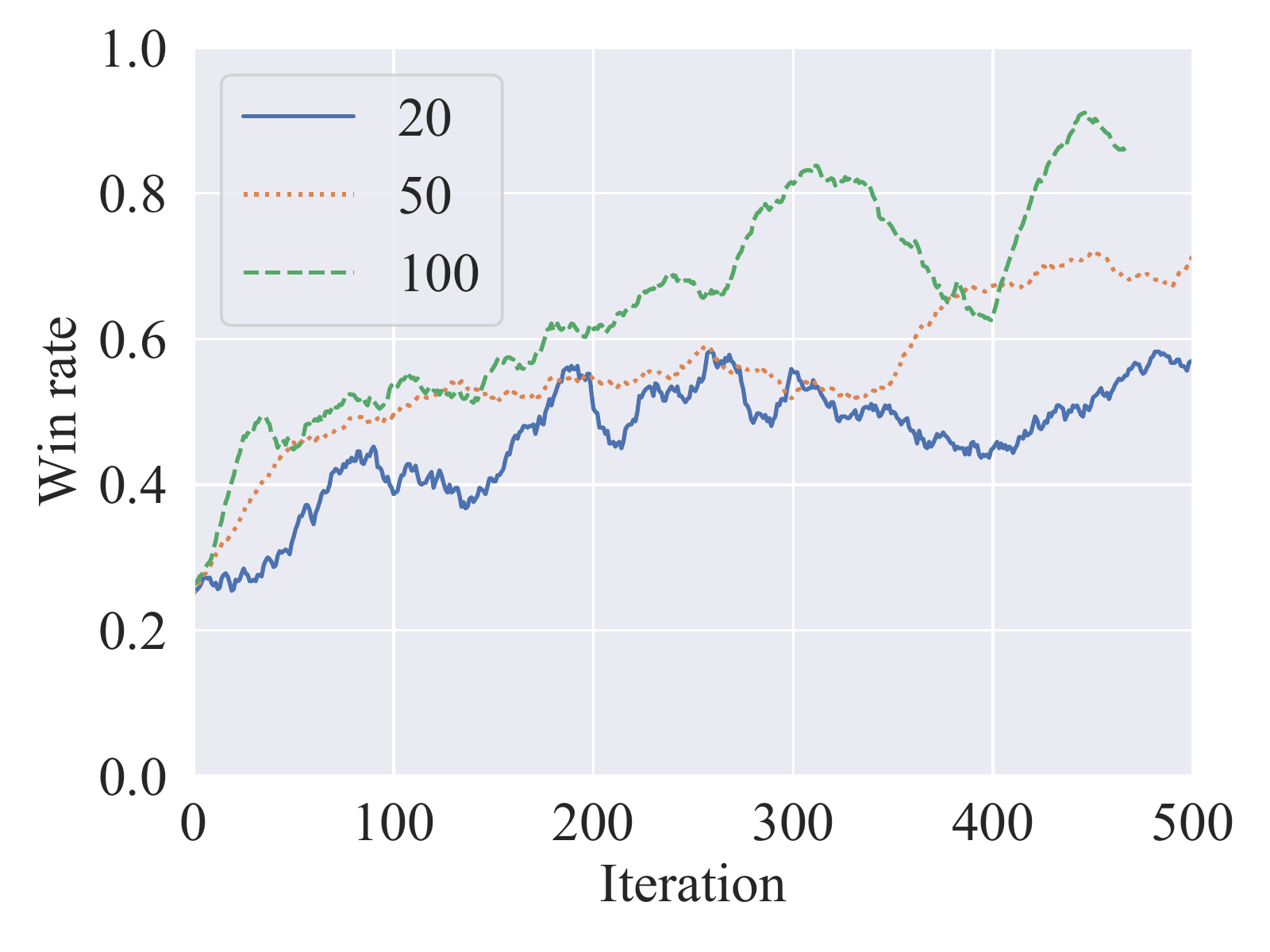}\\
		(d) Impact of episode num
		\label{fig:s.3.2.2}
	\end{minipage}
	\caption{Comparison of settings.}
	\label{fig:s.a.3}
\end{figure*}

\subsection{Comparison of Training Method}
In the following, we will discuss our training processes from three aspects. Firstly, curriculum learning is the major way we use to train our agents against the enemy from low difficulty levels to high difficulty levels. Secondly, our hierarchical architecture can be used for module training, which is convenient for replacing sub-policies and refining some sub-policies while others fixed. Finally, our simultaneous training algorithm is sometimes unstable on high difficulty levels, so we instead use an alternative update strategy.

\subsubsection{Effect of curriculum learning}
In all of our combat models, we firstly train them on low difficulty levels and then transfer them to high difficulty levels. This process follows the way of curriculum learning. We found that when training directly on high difficulty levels, the performance of our agent is difficult to improve. If we start training from low difficulty level and then use the low-difficulty's model as the initial model to update, we could often get better results.

Fig. \ref{fig:s.a.2} (a) shows the comparison between training the pre-trained agent and training from scratch on difficulty level-7. These two agents are using combat network model, and the first agent has been trained on difficulty level-2 and difficulty level-5.

\subsubsection{Effect of module training}
Since we are using a hierarchical architecture, we can easily replace sub-policies. While replacing a sub-policy, the parameters of the controller network and other sub-policies are retained, only the parameters of the newly replaced sub-policy are updated, which can accelerate learning. In the experiment, we have three types of combat strategies. If we have trained the simple combat rule model on high difficulty level, we can use the parameters of its controller network and base network for the other two combat models.

Fig. \ref{fig:s.a.2} (b) shows the comparison between training using pre-trained networks and training from scratch on difficulty level-2. These two agents are using combat network model, and the first agent uses the parameters of controller and base network which are from the combat rule model trained on difficulty level-7.

\subsubsection{Effect of simultaneous updating}
For the sample efficiency, in our training algorithm, we let all the networks of the hierarchical architecture collect their own data and then update at the same time. They are sometimes unstable. Because every network would see the other networks as the part of environment, and do not consider the changes of the others while updating. These unstable training rarely happened on lower difficulty levels, but can be seen on higher difficulty levels. We can use an alternate updating strategy to train the networks which can result in a steady improvement, but need more samples. We can also make the learning rate smaller which could also alleviate this problem. But the training process would be slower and sometimes hard to be improved.

%When performing transfer learning, if all the networks in the hierarchical architecture are updated at the same time, the learning will sometimes become unstable. This may be due to that every network would see the other networks as a stable part of the environment, such will not consider the changes of others. This problem is more often in the higher difficulty levels. Sometimes reducing the learning rate may alleviate this problem, but the cost is that the learning speed may be slower. In this work, we used an alternate updating strategy to train them. This usually leads to a steady improvement at a faster rate.

Fig. \ref{fig:s.a.2} (c) shows the comparison between simultaneously updating and alternately updating on difficulty level-7. The agent uses combat rule model, and has been trained on difficulty level-2 and difficulty level-5.

Fig. \ref{fig:s.a.2} (d) also shows the comparison between simultaneously updating and alternately updating on difficulty level-7, but there are some differences. The agent uses combat network model and has been trained on difficulty level-2 and difficulty level-5. The learning rate is half smaller than Fig. \ref{fig:s.a.2} (c). The green line above is actually loading the trained model of the blue line below. In the experiment, we found that the blue line below could not be improved in the following training period, but the green line above could quickly improve after loading its model.

%The simultaneous training curve (blue line below) is more likely converged and can not be improved in the following. So we use its trained model to be the initial model for the alternative updating (green line above). We can see that the model can still be improved by training them in turn.

\begin{table*}[h]
	% \scriptsize
	% \small
	\centering
	\caption{Evaluation Results.}
	\scalebox{1.0}{
		\begin{tabular}{ l | c c c c c c c | c c c }
			\hline
			Opponent's Type   & \multicolumn{7}{|c|}{Non-cheating (Training)} & \multicolumn{3}{|c}{Cheating (No-training)}  \\
			\hline
			Difficulty Level     & 1 & 2 & 3 & 4 & 5  & 6 & 7 & 8 & 9 & 10  \\
			\hline
			Combat Rule         & 1 & 0.99 & 0.94 & 0.99 & 0.95 & 0.88 & 0.78 & 0.70 & 0.73 & 0.60 \\
			Combat Network   & 0.98& 0.99& 0.45& 0.47& 0.39& 0.73& 0.66& 0.56& 0.52  & 0.41 \\
			Mixture Model       & 1& 1& 0.99& 0.97& 1& 0.90& 0.93& 0.74& 0.71  & 0.43 \\
			\hline
		\end{tabular}
		\label{tab:1}}
	% \vspace{-10pt}
\end{table*}

\subsection{Comparison of Combat Models}
After training the three combat models on difficulty level-7, we perform evaluation for each of them. Each evaluation tests from difficulty level-1 to difficulty level-10. We run $100$ games in each difficulty level and report the winning rate. The result is shown in Table \ref{tab:1}. In the difficulty level-1 to level-7, we found that the training agents have a very good performance. The built-in AI in difficulty level-8 to level-10 uses several different cheat techniques and select different strategies randomly. So the performances of them are unstable and with a bit of randomness. However, it can be seen that our trained agents still have a good performance on fighting against them. The videos of our agents can be searched by "NJU-SC2-Bot" on Youtube.

\subsubsection{Combat rule}
The combat rule agent achieves good results in difficulty level-1 to difficulty level-5. We find that since agents can only use the two basic military units to fight, they are more inclined to use a fast attack fashion (called 'Rush'), which can guarantee the winning rate.

The behaviors of our learned agent is as follow. At the beginning, agents tend to produce more farmers to develop the economy. When the number of farmers is close to saturation, the agent will turn to produce soldiers. When the number of soldiers is sufficient, the agent will choose to attack. This layered progressive strategy is automatically learned through our reinforcement learning algorithm which illustrates the effectiveness of our approach.

\subsubsection{Combat network}
We also test the combat network model on all 10 difficulty levels. We found that although the combat network model achieves a good winning rate on difficulty level-7, the winning rates on several other difficulty levels are not high. It is worth noting that many results of the other levels of difficulty are tie. This means that the model of the combat network is more difficult to completely destroy the enemy (which means eliminating all enemy buildings). Some enemy's buildings or units may be hidden somewhere. If we also see tie as victory, the winning rate of combat network is still high.

%Some enemy's buildings or units maybe hidden somewhere. It should be pointed out that this does not mean that the combat model does not perform well. This is caused by the setting of the victory in the SC2 game. Assuming that we also see tie as victory, the overall winning percentage of the model of the combat network is still high.

%Although the built-in AI in the game can surrender and avoid the tedium of destroying all the enemy buildings, there is no way for agent to accept this surrender, which is caused by the current imperfection of the API provide in SC2LE.

\subsubsection{Mixture model}
It can be found in Table \ref{tab:1} that the mixture model of combat network and combat rule has achieved best results in difficulty level 1-9. This can be explained as follow. The agent can not only choose the attacking area within the camera, but also can switch to a fierce attack on some certain points. This freedom can lead to a performance improvement which makes mixture model best in all three combat models.

%\begin{figure}[h!]
%	\begin{minipage}[t]{\linewidth}
%		\centering
%		\includegraphics[width=\textwidth]{./figure/plot/3.1/3_1.pdf}
%		\caption{Comparison of Combat Model.}
%		\label{fig:3.1}
%		\end{minipage}
%\end{figure}

\subsection{Comparison of Settings}
In this section we have three experiments to show the importance of hierarchy, design of reward and impact of hyper parameters.

\subsubsection{Hierarchy vs non-hierarchy}
As we use a hierarchical architecture, there is a common question which is that whether the SC2 problem can be handled by using a non-hierarchical architecture. In the paper of SC2LE, the learning effect of the original action and state space and non-hierarchical reinforcement learning algorithm has been given. It can be seen that the performance was not satisfied. Our architecture has two levels of abstraction. One is a reduction in the action space, which is done by macro-action. The other is a two-layer architecture that uses the controller to select sub-policies. We tested the effect of keeping the use of macro-actions without using the controller and sub-policies. This is called the single-policy architecture.

We were surprised to find that on low difficulty levels, the single-policy architecture can be almost as good as hierarchical architecture learning which means that macro-actions are effective for training a good agent. However, when moving to high level of difficulty, the final performance of the hierarchical architecture are significantly better than the single-policy architecture, as shown in Fig. \ref{fig:s.a.3} (a), which is trained on difficulty level-7. It can be explained that when the difficulty level is low, the difference between the hierarchical and non-hierarchical architecture is less obvious. When the difficulty of the problem continues to increase, the performance of the hierarchical model will be better than the performance of non-hierarchical model. Another reason for using hierarchy is modularity. Modularity facilitates the replacement of sub-policy. For example, if we need to replace battle sub-policy, we can still retain the parameters of other networks which speeds up training.

%When the agent has learned how to do well in the base, it only needs to change the combat strategy. Modular training facilitates curriculum learning. Therefore, the scalability is also a reason for us to choose this hierarchical framework in SC2.

\subsubsection{Outcome reward vs handcrafted reward}
Win/Loss reward can achieve good results on low difficulty level like handcrafted reward. However, when training is on high difficulty level, we found that the performance of Win/Loss reward on hierarchical model is relatively poor as shown in Fig. \ref{fig:s.a.3} (b) in which the agent is using combat network model and trained on difficulty level-7.

In SC2LE, it is mentioned that one can use Blizzard score as a reward. However, we found that Blizzard score and winning rate are sometimes not in a proportional relationship, especially on low difficulty levels. While the agent is trying to improve the score, it may ignore the attack chance on the enemy base and lose the best opportunity to win. This is shown in Fig. \ref{fig:s.a.3} (c) in which the agent is using combat network model and trained on difficulty level-2.

%The final performance of the agent trained in SC2LE by Blizzard score also shows this problem. That is the reason why Blizzard score is not an ideal reward function.

\subsubsection{Influence of hyper-parameters}
We have experimented with a variety of different parameters and found that the number of episodes in each iteration has some influence on the learning speed of the agent. When the number is small, the learning will be unstable and hard to converge. Improving the number of episodes can mitigate the problem. This effect is shown in Fig. \ref{fig:s.a.3} (d) in which the agent is using combat rule model and trained on difficulty level-2. Other parameters have little effect on training.

%\subsection{Discussion}
%Strategies used by built-in AIs are highly optimized by game developers and therefore pose significant challenges to learning algorithms. Our method is characterized by the abstraction and reduction of StarCraft II from multiple levels. For example, by learning the macro-actions, the action space in the StarCraft II is greatly reduced. Long-time horizon problem of the StarCraft II is mitigated through a hierarchical approach. This learning algorithm is robust (because of the characteristics of learning) and efficient (because of the abstraction of problems). The combination of these factors makes this framework suitable for solving large-scale reinforcement learning problem such as StarCraft II. Our current approach still has some shortcomings. For example, the 64x64 map we are currently testing is small, we only use the two arms of the initial level. In the future, we will explore learning on larger maps and try to use more arms to organize tactics.

\section{Conclusion}
In this paper, we investigate a set of techniques of reinforcement learning for the full-length games in StarCraft II, including hierarchical policy training with extracted macro-actions, reward design, and curriculum design. With limited computation resource, we show that the combined approach achieves the state-of-the-art results beating the built-in AI. In the future, we will continue to explore more RL algorithms that can better solve large-scale problems.

% \vskip 0.2in
\bibliographystyle{abbrv}
\bibliography{reference}

\newpage
\appendix
\section{Appendix}
\subsection{State space}
In our hierarchical architecture, the states has two types, which are the non-spatial features and the spatial features. The list of the non-spatial features is shown in Table \ref{tab:2}. The content of the spatial features is based on Table \ref{tab:3}. The controller only uses some of the non-spatial features as its global state. The base policy uses all the non-spatial features as its local state. The battle policy use both the non-spatial and spatial features as its local state. 

\begin{table*}[h]
	% \scriptsize
	% \small
	\centering
	\caption{Non-spatial features}
	\scalebox{1.0}{
		\begin{tabular}{ l | l }
			\hline
			features   &  remarks  \\
			\hline
			opponent.difficulty  & from 1 to 10 \\
			observation.game-loop & game time in frames  \\
			observation.player-common.minerals & minerals  \\
			observation.player-common.vespene & gas \\
			observation.score.score-details.spent-minerals & mineral cost \\
			observation.score.score-details.spent-vespene  & gas cost \\
			player-common.food-cap & max population \\
			player-common.food-used & used population \\
			player-common.food-army  & population of army \\
			player-common.food-workers  & population of workers(probes) \\
			player-common.army-count & counts of army \\
			player-common.food-army / max(player-common.food-workers, 1) & rate of army on workers \\
			num of army & the number of our army \\
			* num of probe, zealot, stalker, etc & multi-features \\
			* num of pylon, assimilator, gateway, cyber, etc &  multi-features \\
			* score\_cumulative & blizzard detailed score(multi-features) \\
			\hline
			* cost of pylon, assimilator, gateway, cyber, etc &  multi-features \\
			* num of probe which are doing building actions &  multi-features \\
			\hline
			* num of probe for mineral and the ideal num & multi-features \\
			* num of probe for gas and the ideal num & multi-features \\
			* num of the training probe, zealot, stalker &  multi-features \\
		\end{tabular}
		\label{tab:2}}
	% \vspace{-10pt}
\end{table*}

\begin{table*}[h]
	% \scriptsize
	% \small
	\centering
	\caption{spatial features}
	\scalebox{1.0}{
		\begin{tabular}{ l | l }
			\hline
			features   &  post processing (map-width is set to 64)  \\
			\hline
			observation(minimap)[height] & reshape(-1, map-width, map-width) / 255 \\
			observation(minimap)[visibility] & reshape(-1, map-width, map-width) / 2 \\
			observation(minimap)[camera]& reshape(-1, map-width, map-width) \\
			observation(minimap)[relative]& reshape(-1, map-width, map-width) / 4 \\
			observation(minimap)[selected]& reshape(-1, map-width, map-width)\\
			
			observation(screen)[relative]& reshape(-1, map-width, map-width) / 4 \\
			observation(screen)[selected]& reshape(-1, map-width, map-width) \\
			observation(screen)[hitpoint-r]& reshape(-1, map-width, map-width) / 255 \\
			observation(screen)[shield-r]& reshape(-1, map-width, map-width) / 255 \\
			observation(screen)[density-a]& reshape(-1, map-width, map-width) / 255 \\
		\end{tabular}
		\label{tab:3}}
	% \vspace{-10pt}
\end{table*}

\subsection{Macro actions}
%Here is how we generate macro actions. First we let the experts play 36 games including difficult level-1 to level3. In these games, we use the Protoss against the Terran, and only use the first two arms. After that, we saved the replays of the experts and analyzed these replays. The sequence of actions we dig through the PrefixSpan algorithm is as follows. The top 30 action sequences with the highest frequency of occurrence are listed in Table \ref{tab:macro-action}. It is worth noting that since all StarCraft 2 operations follow a form similar to English grammar, that is, the form of the subject plus verbs, the first action of all action sequences is necessarily the selection action. In addition, in the action sequence, many actions are smart screen operations. This operation produces different effects depending on the target of execution. Therefore, it is not helpful for building macro actions, and can be discarded. When we have dropped some duplicate action sequences, we can construct a collection of macro actions with the remaining ones. It should be noted that since macro actions are not code, we need to "translate" them into specific code. These codes need to be carefully written to ensure that they perform the sequence of operations in macro actions.

The Macro actions are generated by using the PrefixSpan algorithm on our expert's game replays. Table \ref{tab:macro-action} shows the top 30 action sequences with the highest frequency of occurrence. As we can see, the first action of the action sequences is necessarily the selection action. In addition, many actions in the action sequences are about screen movement. This operation produces different effects depending on the target of execution. Therefore, it is not helpful for building macro actions and can be discarded. When we have dropped some duplicate or meaningless action sequences, we can construct a collection of macro actions with the remaining ones. It should be noted that since macro actions are not code, we need to "translate" them into specific code. These codes need to be carefully written to ensure that they perform the sequence of operations in macro actions.

\begin{table*}[h]
	% \scriptsize
 \small
	\centering
	\caption{Prefixspan-Result (30 most frequently occurring)}
	\scalebox{1.0}{
		\begin{tabular}{ l  l  c }
			\hline
			action sequences   & frequency  \\
			\hline
			select-point(unit name: Protoss.Probe) $\to$ Harvest-Gather-Probe-screen & 2711 \\
			select-point(unit name: Protoss.Probe) $\to$ Smart-screen & 2253  \\
			select-point(unit name: Protoss.Nexus) $\to$ Train-Probe-quick & 1421  \\
			select-point(unit name: Protoss.Probe) $\to$ Build-Pylon-screen & 1298 \\
			select-point(unit name: Protoss.Nexus) $\to$ Smart-screen & 1030 \\
			select-point(unit name: Protoss.Nexus) $\to$ select-army & 968 \\
			select-point(unit name: Protoss.Probe) $\to$ Build-Gateway-screen & 814 \\
			select-point(unit name: Protoss.Gateway) $\to$ Train-Zealot-quick & 762 \\
			select-point(unit name: Protoss.Probe) $\to$ Build-Pylon-screen $\to$ Harvest-Gather-Probe-screen & 659 \\
			select-point(unit name: Protoss.Gateway) $\to$ Train-Stalker-quick & 621 \\
			select-point(unit name: Protoss.Nexus) $\to$ select-army $\to$ Attack-Attack-screen & 581 \\
			select-point(unit name: Protoss.Probe) $\to$ select-army & 574 \\
			select-point(unit name: Protoss.Nexus) $\to$ Train-Zealot-quick & 536 \\
			select-point(unit name: Protoss.Gateway) $\to$ select-army & 467 \\
			select-point(unit name: Protoss.Probe) $\to$ Build-Gateway-screen $\to$ Harvest-Gather-Probe-screen & 466 \\
			select-point(unit name: Protoss.Nexus) $\to$ Train-Stalker-quick & 325 \\
			select-point(unit name: Protoss.Gateway) $\to$ Smart-screen & 300 \\
			select-point(unit name: Protoss.Gateway) $\to$ select-army $\to$ Attack-Attack-screen & 284 \\
			select-point(unit name: Protoss.Probe) $\to$ Attack-Attack-screen & 260 \\
			select-point(unit name: Protoss.Probe) $\to$ Build-Pylon-screen $\to$ Smart-screen & 258 \\
			select-point(unit name: Protoss.Probe) $\to$ select-army $\to$ Attack-Attack-screen & 254 \\
			select-point(unit name: Protoss.Probe) $\to$ Build-Assimilator-screen & 253 \\
			select-point(unit name: Protoss.Probe) $\to$ Train-Zealot-quick & 243 \\
			select-point(unit name: Protoss.Gateway) $\to$ Cancel-BuildInProgress-quick & 209 \\
			select-point(unit name: Protoss.Probe) $\to$ Smart-screen $\to$ Build-Pylon-screen & 194 \\
			select-point(unit name: Protoss.Probe) $\to$ Build-CyberneticsCore-screen & 182 \\
			select-point(unit name: Protoss.Probe) $\to$ Smart-screen $\to$ select-army & 166 \\
			select-point(unit name: Protoss.Probe) $\to$ Smart-screen $\to$ Harvest-Gather-Probe-screen & 164 \\
			select-point(unit name: Protoss.Probe) $\to$ Smart-screen $\to$ Build-Gateway-screen & 147 \\
			select-point(unit name: Protoss.Probe) $\to$ Harvest-Gather-Probe-screen $\to$ select-army & 147 \\
		\end{tabular}
		\label{tab:macro-action}}
	% \vspace{-10pt}
\end{table*}

\subsection{Reward Function}
The reward function has two type. One is dense or instant reward, which can be got after every action. The other is sparse or final reward, which only can be got in the terminal state. The controller and the sub-policies use the same final reward, which is the win/lose and the time penalty. For each sub-policy, the dense or instant reward is different. In our best version, the instant rewards depend on the number of the units and structures. For example, the number of probe, pylon, army and so on. There is also another version which performed slightly worse than before. It uses the Blizzard detailed score as the reward. More specifically, the base policy use the worker idle time, the production idle time, total value unit and total value structures as its instant reward. The battle policy use the killed value units and killed value structures as its reward. The instant reward for the controller is the cumulative sum of the chosen sub-policy's instant reward in the time interval.

\subsection{Training Setting}
In the experiment, we used a machine with 4 GPUs and 48 CPUs. There were 10 workers and each worker had 5 threads. Every thread was corresponding to a SC2 environment. In other words, There were 50 running environments at the same time.  They all shared the global hierarchical architecture and also the parameters of networks. Since we used PPO algorithm for the training, our distributed system was simultaneous. In each iteration, we made the workers run 100 full-length games of SC2 and each worker would collect the data from 10 episodes. It would takes about 4 minutes. When all the workers had finished collecting the data, each worker would compute the gradients on its own. Finally, all the gradients from the wokers would be gathered and used to update the global hierarchical architecture.

\subsection{Building Location}
It is worth noting that the construction of buildings had to be given a location. There are mainly three ways to get it. The first one is by learning, the second one is by heuristic algorithms or human knowledge, and the third one is by random. The method by learning or human knowledge is too expensive, and most locations had little influence on our experiment. Therefore, after some trials, we chosed the location by random.

\begin{figure}[h!]
	\begin{minipage}[t]{\linewidth}
		\centering
		\includegraphics[width=\textwidth]{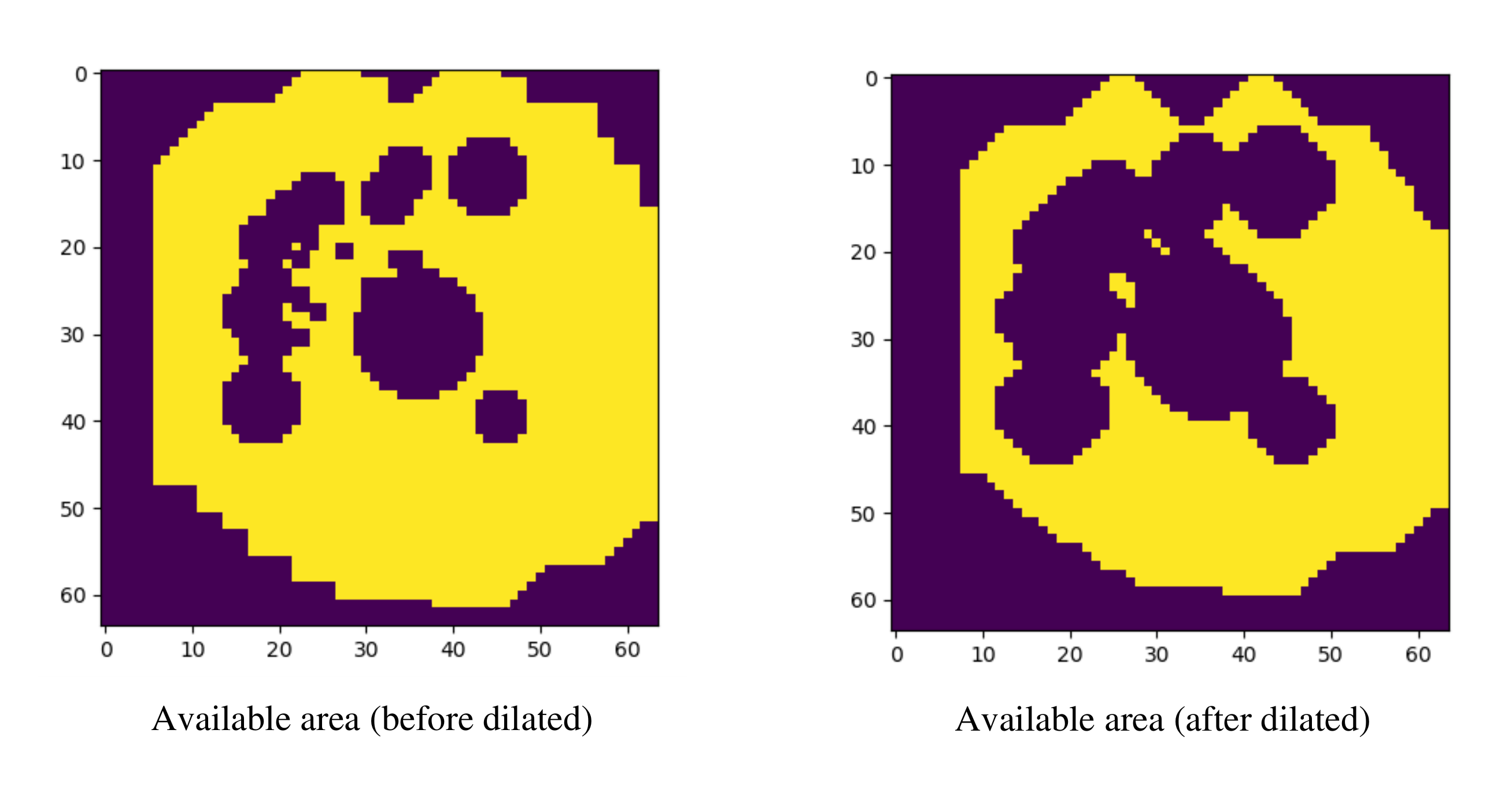}
		\caption{Effect of dilation algorithm.}
		\label{fig:build}
	\end{minipage}
\end{figure}

However, random selection without restriction would cause conflicts with other buildings or units, since the building that was going to build havd its collision volumes. We selected a commonly used dilation algorithm in the field of image processing research to solve this problem. This method would reduce the available area depended on the size of the building which was going to build. The success rate of the construction operation on the reduced area was greatly improved. In Fig. \ref{fig:build}, the left figure is an original available map, and the right one is a reduced map after the dilating operator. It can be seen that the area that can be constructed (yellow) becomes smaller after dilation algorithm. Protoss is a relatively special race that requires the supply of power in addition to the need for open space. Therefore, when building a building other than Pylon, the area has to be logically AND with a power mask screen to obtain the area map that can be finally constructed.

\end{document}